\documentclass[]{interact}
\usepackage{epstopdf}
\usepackage{subfigure}
\usepackage{tabularx}

\usepackage{natbib}
\bibpunct[, ]{(}{)}{;}{a}{}{,}
\renewcommand\bibfont{\fontsize{10}{12}\selectfont}
\usepackage{hyperref}

\usepackage{listings}
\usepackage{multirow}
\usepackage{float}

\theoremstyle{plain}

\theoremstyle{definition}

\theoremstyle{remark}

\begin{document}

\title{Evaluating Large Language Models for Financial Reasoning: A CFA-Based Benchmark Study}

\author{
\name{Xuan Yao, Qianteng Wang, Xinbo Liu, Ke-Wei Huang}
}
\maketitle
\makeatletter
\renewcommand{\history}{}
\makeatother
\begin{abstract}
The rapid advancement of large language models (LLMs) presents significant opportunities for financial applications, yet systematic evaluation of their capabilities in specialized financial contexts remains limited. While existing studies have explored general financial NLP tasks, the unique challenges of professional financial certification, including long contexts, complex quantitative reasoning, and domain-specific knowledge requirements demand rigorous assessment before real-world deployment.

This study presents the first comprehensive evaluation of state-of-the-art large language models (LLMs) distinguished by their core design priorities using 1,560 multiple-choice questions from official CFA mock exams across Levels I–III. Specifically, we compare the performance of models that are (i) multimodal and computationally powerful, (ii) specialize in reasoning and highly accurate, and (iii) lightweight and efficiency-optimized. Unlike previous studies that focused on general financial NLP tasks, CFA examinations present unique challenges including progressively complex case-based scenarios with lengthy contexts (averaging 2,500+ characters at Level III), multiple integrated tables, and demands for both conceptual mastery and quantitative precision across ten specialized financial domains. As one of the most rigorous professional certifications globally, CFA provides an authentic benchmark that mirrors real-world financial analysis complexity. We assess models under zero-shot prompting conditions and through a novel domain reasoning Retrieval-Augmented Generation (RAG) pipeline that integrates official CFA curriculum content. The RAG system achieves precise domain-specific knowledge retrieval and effective context integration through hierarchical knowledge organization and structured query generation, significantly enhancing large language model reasoning accuracy in professional financial certification evaluation. 

Results reveal that reasoning and accuracy oriented models consistently outperforms other models with zero-shot accuracies, while the RAG pipeline provides substantial improvements particularly for complex scenarios. Comprehensive error analysis identifies knowledge gaps as the primary failure mode, with minimal impact from text readability.

These findings provide actionable insights into LLM deployment in finance, providing practitioners with evidence-based guidance for model selection and cost-performance optimization.

\end{abstract}


\section{Introduction}

\subsection{Background and Motivation}

The field of large language models (LLMs) has brought transformative changes across a spectrum of reasoning and cognitive tasks. Models such as OpenAI's GPT series, Google's Gemini, Meta's LLaMa and Anthropic's Claude are demonstrating unprecedented capabilities in natural language understanding, generation, and deduction. These advances have pushed the boundaries of state-of-the-art in universal domains, ranging from conversational AI, programming assistance to education and scientific research.

Among the various sectors undergoing rapid transformation through the integration of artificial intelligence, the financial domain emerges as particularly promising. The financial system presents distinct linguistic and analytical challenges, encompassing a broad range of tasks such as market forecasting, risk assessment, algorithmic trading, regulatory interpretation, and client interaction. These tasks necessitate the extraction and interpretation of unstructured textual data such as financial statements, news articles, analyst commentaries, and regulatory filings, in conjunction with advanced quantitative reasoning over structured financial datasets.

Recent LLM research has been consistently trying to bridge this gap, showing expanding promise in tasks such as analyzing financial sentiments, generating synthetic datasets, providing investment advice, and predicting market trends \citep{Lee2024}. Notably, newer LLM applications incorporate techniques such as Retrieval-Augmented Generation (RAG), fine-tuning on domain-specific corpora, and integration with various forms of knowledge base, all of which demonstrate performance improvement on specialized tasks in finance. 

Parallel to the rapid evolution and proliferation of LLMs, it remains a challenge to systematically assess their capabilities within the financial domain. Given the high potential and domain-specific complexity in finance, a rigorous understanding of the strengths and limitations of LLMs is essential before integrating them into real-world applications. For practitioners aiming to harness these models effectively, such fundamental evaluation is not only beneficial but imperative.

Meanwhile, many studies in the intersection of LLM and finance have investigated general financial tasks rather than structured exams. Unlike previous studies that examined isolated tasks such as sentiment analysis or named entity recognition, CFA examinations mirror the integrated complexity of professional financial practice, making it an ideal proxy for evaluating whether LLMs can perform at the level expected in consequential financial decision-making contexts.

\subsection{Research Objectives and Approach}
This study seeks to comprehensively investigate LLMs' performance in finance, highlighting their capabilities, failure modes, and other emergent behaviors. Beyond benchmarking, we address their current limitations through prompt engineering strategies, specifically by introducing a domain reasoning RAG pipeline. This pipeline enhances contextual comprehension and domain-specific reasoning by integrating external financial materials into the LLM workflow. 

The \cite{CFAInstitute} program offers a uniquely rigorous benchmark that addresses critical limitations in existing LLM financial evaluations. CFA examinations require simultaneous mastery of quantitative analysis, qualitative reasoning, ethical decision-making, and comprehensive domain knowledge within individual questions. According to the \cite{cfastructure}, the program's three-tiered structure reflects authentic career progression in finance—from foundational concepts (Level I) for entry-level positions in finance like Investment Banking Analyst through intermediate application (Level II) for positions like Equity Research Analyst to expert-level synthesis and portfolio management (Level III) for senior positions like Portfolio Manager. This professional authenticity enables our evaluation to be distinguished from laboratory-style assessments and provides direct insights into LLM readiness for deployment in investment management, corporate finance, and advisory roles where analytical errors carry significant financial and regulatory consequences.

In this study, we compare the performance of models that are (i) multimodal and com-
putationally powerful, (ii) specialize in reasoning and highly accurate, and (iii)
lightweight and efficiency-optimized. Three benchmark representative LLMs are \texttt{GPT-4o}, \texttt{GPT-o1}, and \texttt{o3-mini}.  \texttt{GPT-4o} is OpenAI’s generalist flagship model, engineered to reason fluidly across text, vision, and audio in real time (\cite{gpt-4o}). It maintains GPT-4–level accuracy while offering dramatically lower latency and cost. Unlike speed-optimized GPT variants, o1 is architected to spend extra compute on extended chains-of-thought before emitting an answer, which often boosts logical consistency in complex multi-step problems (\cite{gpt-o1}). That additional reasoning depth comes at the cost of higher inference time and compute demand, but GPT-o1 provides a significant enhancement for complicated reasoning queries and brings a new level of AI capability. o3-mini represents the newest generation of small-footprint reasoning models (\cite{o3-mini}). Despite its reduced parameter count, it delivers strong performance while remaining price-efficient for large-scale batch evaluation. Its rapid responses and favorable cost-latency profile make it ideal for high-volume tasks.

Taken together, these three models offer a representative snapshot of the current LLM landscape: \texttt{GPT-4o} as a versatile multimodal flagship with strong general capabilities, \texttt{GPT-o1} as a cutting-edge model in 2024 optimized for complex reasoning tasks, and \texttt{o3-mini} as a lightweight, cost-effective model ideal for high-volume or latency-sensitive applications. While the performance gains between successive models are increasingly incremental rather than revolutionary, this trio provides a robust and timely proxy for the range of capabilities practitioners can expect across the product cycle in the near term.

Our methodology employs two complementary evaluation approaches that systematically assess different dimensions of LLM financial competency.

First, we establish baseline capabilities through Zero-Shot Baseline evaluation, which reveals models' intrinsic financial knowledge without external assistance. This approach is critical for understanding what financial concepts, formulas, and reasoning patterns LLMs have internalized during pre-training, providing insights into their foundational readiness for financial applications. Zero-shot performance indicates whether models can independently handle routine financial analysis tasks or require augmentation for professional deployment.

Second, we implement a customized Domain Reasoning RAG Pipeline that addresses the fundamental challenge of domain-specific knowledge gaps identified in zero-shot evaluation. This pipeline specifically targets the knowledge-intensive nature of professional finance by integrating official CFA curriculum materials—the authoritative source documents that define industry standards. Our RAG approach employs a novel two-stage process: (1) Query Generation, where models analyze questions to produce targeted summaries and keyword lists, ensuring retrieval precision rather than generic document search; and (2) Context-Augmented Reasoning, where models leverage retrieved curriculum content to enhance their responses. 

This dual-methodology framework provides a comprehensive assessment of LLM capabilities across the spectrum of financial reasoning requirements. Zero-shot evaluation reveals intrinsic strengths and limitations, while RAG evaluation demonstrates how effectively models can integrate external knowledge sources—a critical capability for real-world financial applications where practitioners must synthesize information from multiple authoritative sources including regulations, market data, and institutional guidelines. Together, these approaches enable us to distinguish between fundamental reasoning limitations and knowledge accessibility challenges, providing actionable guidance for practitioners on when and how to deploy LLMs in financial contexts. The adaptability of our framework allows for seamless incorporation of proprietary data sources, making it a valuable blueprint for deploying LLMs in various real-world financial environments.

Additionally, this work aims to serve as a foundational guide for financial professionals and institutions considering LLM integration. By benchmarking model performance across diverse topics and long contexts, we reveal nuanced strengths and potential pitfalls that practitioners should account for. Furthermore, the proposed RAG pipeline demonstrates how LLMs can be enhanced through targeted contextualization, with potential applications in analyzing financial statements, interpreting regulatory documentation, or supporting client advisory functions. 

\section{Literature Review}

Financial LLM evaluation research has progressed from broad task-based assessments to specialized evaluation methodologies, ultimately converging on targeted technical solutions for identified challenges.

Initial research established the foundational understanding of LLM capabilities across diverse financial contexts. \cite{li2023chatgptgpt4generalpurpose} conducted the earliest comprehensive assessment spanning sentiment analysis, named entity recognition, relation extraction, and question answering in financial contexts. Their findings revealed a fundamental pattern: while LLMs excel in language understanding tasks, they struggle significantly with domain-specific knowledge application. This seminal work established that financial competency extends beyond isolated task performance, requiring integrated reasoning capabilities across multiple domains.

Building on these foundational insights, researchers developed more rigorous evaluation frameworks through standardized financial certification exams. \cite{callanan2023gptmodelsfinancialanalysts} pioneered the use of CFA level I and II mock exams to assess ChatGPT and GPT-4, employing zero-shot, chain-of-thought (CoT), and few-shot prompting strategies. Their comprehensive error taxonomy—spanning knowledge, reasoning, calculation, and inconsistency errors—provided a systematic framework for understanding the specific limitations identified in the broader task evaluations. \cite{Fairhurst02012025} expanded this certification-based approach across FINRA and NASAA licensing exams, though these assessments primarily evaluate regulatory compliance and client communication rather than the analytical rigor central to financial reasoning.

Recognizing the limitations of both broad task assessments and basic certification approaches, researchers began developing enhanced methodologies to address identified gaps. \cite{mahfouz-etal-2024-state} extended evaluation across all CFA levels while introducing RAG methodologies, demonstrating the potential of combining CoT reasoning with retrieval augmentation. However, their reliance on pre-processed textbook content highlighted a critical limitation: the need to handle complex, unstructured documents characteristic of real-world financial analysis. This gap became particularly evident when contrasted with the integrated, multi-competency reasoning that CFA examinations demand—precisely the holistic capability that \cite{li2023chatgptgpt4generalpurpose} had identified as lacking in isolated task performance.

These comprehensive evaluations converged on two fundamental technical challenges that shape current research directions. \cite{srivastava2024evaluatingllmsmathematical} identified the mathematical reasoning bottleneck: effective financial analysis demands seamless integration of textual comprehension and numerical computation, areas where the domain-specific knowledge limitations first observed by \cite{li2023chatgptgpt4generalpurpose} become particularly pronounced. The document processing challenge emerged as equally critical—financial reasoning requires accurate extraction and preservation of quantitative information from complex, semi-structured documents.
Recent research has focused on developing targeted technical solutions to address these identified challenges. \cite{wang2024mineruopensourcesolution} introduced MinerU, a document processing framework that preserves critical formatting elements including tables and mathematical expressions during PDF-to-markdown conversion. This advancement directly addresses the document processing bottleneck identified throughout the evaluation research, enabling more effective RAG applications where preservation of quantitative information and structured data is essential for the integrated reasoning capabilities that financial analysis demands.

Despite these advances, several limitations persist in current research. Most studies focus on isolated aspects of financial reasoning—either exam performance, specific NLP tasks, or mathematical computation—without comprehensive integration. Additionally, the temporal dynamics of financial knowledge, where regulations and market practices evolve continuously, remain underexplored in existing evaluation frameworks. The gap between laboratory-style assessments and real-world deployment considerations, including cost-performance trade-offs and error tolerance requirements, suggests opportunities for more applied research in financial LLM applications. 

\section{Data and Methodology}

\subsection{Data}

The CFA examinations provide a rigorous, globally recognized system for assessing the financial analysis capabilities of LLMs. Candidates must successfully pass three levels of exams, hold a bachelor's degree, and complete three years of professional financial experience to earn the charter designation. Each level of the curriculum and exam builds on the previous level and becomes increasingly complex, according to \cite{CFAInstitute}. Level I exam features short multiple-choice questions (MCQs) with tables in some questions and tests fundamental knowledge of the investment industry. In contrast, both the Level II and III exams include a case description with several paragraphs and multiple tables in the majority of cases. Level II exam uses MCQ to test candidates' ability to leverage accumulated knowledge and analyze various scenarios. In Level III, a mix of MCQs and short essay questions are presented to assess how candidates integrate a wide range of concepts and apply them to real-world situations.

We collected a dataset of the most recent mock exams offered by the CFA Institute, which closely replicate the format and complexity of the actual exams. It includes multiple-choice questions (MCQs) from five 2024 Level I mock exams, five 2025 Level II mock exams, and five Level III mock exams from 2022 to 2024. Specifically, each Level I, Level II, and Level III mock exam comprises 180, 88, and 44 MCQs, respectively. Table \ref{table:CFA_topics} presents an overview of the topic distribution across the mock exam questions. Among the ten topics appeared in Level I and II, seven topics are explicitly tested in Level III. However, the remaining three topics are implicitly integrated into the Portfolio Management topic. Table \ref{tab:question_lenth} displays the average character length of the MCQs. Consequently, this dataset challenges the model to comprehend lengthy case contexts, extract pertinent information from both tables and text, and select the correct answer from three possible options.

The CFA Institute uses a minimum passing score (MPS) to determine whether a candidate can pass the exam. The MPS is obtained through a standard internal procedure and set by the Board of Governors. It is based on a criterion-referenced model which represents a specific standard of performance required. The percentage of questions a candidate needs to answer correctly is based on the difficulty of the version of the exam taken, as the questions in one exam are randomly drawn from a question pool. Therefore, the percentage of correct answers required to pass the exam is not disclosed to the public and may vary slightly in different question sets. Since we utilize a large set of questions, we employ an estimated passing criteria proposed by \cite{callanan2023gptmodelsfinancialanalysts}, which are based on discussions on online forums:
\begin{enumerate}
    \item Level I - achieve an accuracy of at least 60\% in each topic and an overall accuracy of at least 70\%
    \item Level II and III - achieve a score of at least 50\% in each topic and an overall score of at least 60\%
\end{enumerate}

\begin{table}[ht]
\centering
\small
\begin{tabular}{lcccc}
\hline
\textbf{Topic} & \textbf{Level 1} & \textbf{Level 2} & \textbf{Level 3} & \textbf{Total} \\
\hline
Alternative Investments & 65 (7.2\%) & 32 (7.3\%) & 4 (1.8\%) & 101 (6.5\%) \\
Corporate Issuers & 69 (7.7\%) & 28 (6.4\%) & - & 97 (6.2\%) \\
Derivatives & 70 (7.8\%) & 28 (6.4\%) & 20 (9.1\%) & 118 (7.6\%) \\
Economics & 66 (7.3\%) & 28 (6.4\%) & 22 (10.0\%) & 116 (7.4\%) \\
Equity Investments & 115 (12.8\%) & 64 (14.5\%) & 16 (7.3\%) & 195 (12.5\%) \\
Ethics & 135 (15.0\%) & 60 (13.6\%) & 42 (19.1\%) & 237 (15.2\%) \\
Fixed Income & 110 (12.2\%) & 56 (12.7\%) & 36 (16.4\%) & 202 (12.9\%) \\
Financial Statement Analysis & 110 (12.2\%) & 60 (13.6\%) & - & 170 (10.9\%) \\
Portfolio Management & 90 (10.0\%) & 56 (12.7\%) & 80 (36.4\%) & 226 (14.5\%) \\
Quantitative Methods & 70 (7.8\%) & 28 (6.4\%) & - & 98 (6.3\%) \\
\hline
\textbf{Total} & 900 (100.0\%) & 440 (100.0\%) & 220 (100.0\%) & 1560 (100.0\%) \\
\hline
\end{tabular}
\caption{CFA Topics Distribution by Level (Count and Percentage)}
\label{table:CFA_topics}
\end{table}

\begin{table}[H]
\centering
\small
\begin{tabular}{lccc}
\toprule
\textbf{Topic} & \textbf{Level 1} & \textbf{Level 2} & \textbf{Level 3} \\
\midrule
Alternative Investments  & 112 & 2144 & 1941 \\
Corporate Issuers  & 114 & 1970 & -  \\
Derivatives  & 107 & 1945 & 3296 \\
Economics & 92  & 1989 & 2732 \\
Equity Investments  & 129 & 2076 & 2885 \\
Ethics & 344 & 2536 & 2674 \\
Fixed Income  & 139 & 1794 & 2059 \\
Financial Statement Analysis & 149 & 2253 & - \\
Portfolio Management  & 132 & 2021 & 2873\\
Quantitative Methods  & 138 & 2757 & - \\
\midrule
\textbf{Total} & \textbf{146} & \textbf{2148} & \textbf{2538} \\
\bottomrule
\end{tabular}
\caption{Average Question Length in CFA Exam Questions (in Characters)}
\label{tab:question_lenth}
\end{table}

\subsection{Evaluation Approaches}

\subsubsection{Zero-Shot Baseline}

The zero-shot prompt combines with instruction prompting to assess the models' intrinsic capabilities. The system prompt used is as follows.

\begin{lstlisting}

Assume you are taking a CFA (Chartered Financial Analyst) Level {level} exam. 
You will be given a question and three possible answers (A, B, or C). Your task is to choose the correct answer and provide a brief explanation. 
Follow these steps to solve the question:
1. Read and understand the question carefully.
2. Identify applicable CFA curriculum concepts, such as financial theories, principles, or formulas.
3. Analyze each option using the facts from the question and the key concepts.
4. Choose the best answer and provide a brief explanation within 100 words.

Respond in this format:  
Choice: [A, B, or C]  
Explanation: [Briefly explain your choice within 100 words.]

\end{lstlisting}

\subsubsection{domain RAG Pipeline}

In this section, we evaluate LLMs' capabilities in utilizing external contexts using a domain reasoning RAG pipeline. We leverage the official CFA curriculum from the corresponding year of the mock exam. The CFA curriculum is the basis for the exams and covers all topics in detail. Hence, it poses a challenge for LLMs to locate relevant snippets and decide whether the snippets are useful. In addition, the source PDF files are converted into markdown using MinerU \cite{wang2024mineruopensourcesolutionprecise} for better representation of formulas and tables.

\begin{figure}[h]
    \centering
    \includegraphics[width=0.5\textwidth]{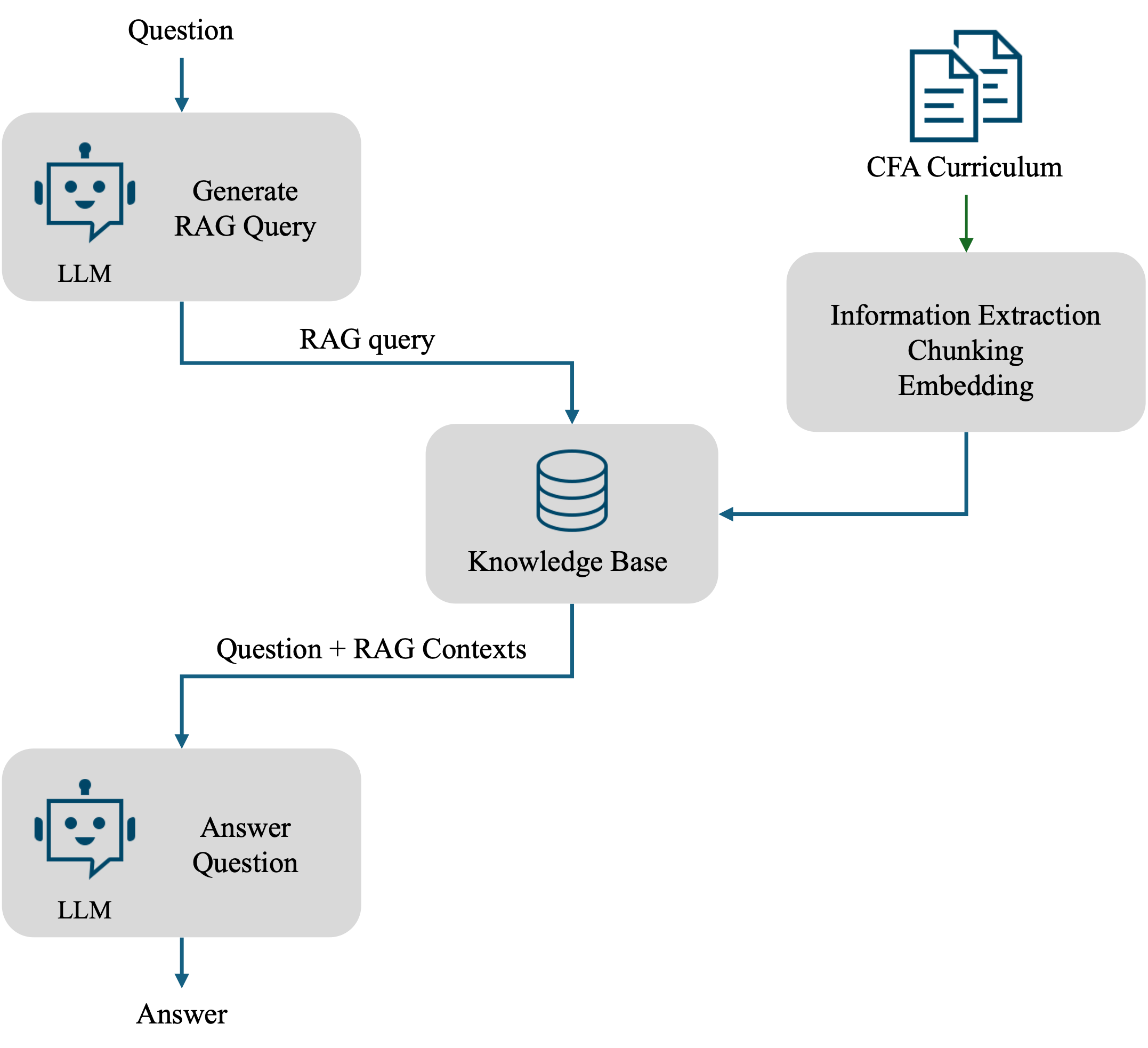}
    \caption{Domain Reasoning RAG Pipeline}
    \label{fig:rag_pipeline}
\end{figure}

The pipeline is illustrated in Figure \ref{fig:rag_pipeline} and summarized as follows. 
\begin{enumerate}
    \item \textbf{Generate RAG query:} The model is fed with the question and options, and it is asked to summarize what the question is asking for and give a list of keywords. The prompt used is as follows:
    \begin{lstlisting}
You are a CFA (Chartered Financial Analyst) Level {level} candidate.  
You will be given a CFA exam question.  
Your task is to:  
1. Summarize what the question is asking for in clear, concise language (no more than 50 words).  
2. Identify 5-10 relevant keywords related to financial concepts or CFA curriculum topics that are essential to solving the question.

Format your response as follows:  
Summary: [Your concise summary within 50 words]  
Keywords: [List of 5-10 essential keywords]
    \end{lstlisting}
    \item \textbf{Retrieve contexts:} The RAG query generated by the model is used to extract five pieces of references from the CFA curriculum of the corresponding topic.
    \item \textbf{Solve the question:} The question is augmented with the RAG references and presented to the LLM to obtain an answer and a brief explanation, using the following system prompt.
    \begin{lstlisting}
You will be given a question and three possible answers (A, B, or C). You will also be given several references from the CFA curriculum.
Your task is to choose the correct answer and provide a brief explanation. 
Follow these steps to solve the question:
1. Read and understand the question carefully.
2. Identify applicable CFA curriculum concepts, such as financial theories, principles, or formulas.
3. Analyze each option using the facts from the question and the key concepts.
4. Choose the best answer and provide a brief explanation within 100 words.

Respond in this format: 
Choice: [A, B, or C]
Explanation: [Briefly explain your choice within 100 words.]
    \end{lstlisting}
\end{enumerate}

\subsection{RAG Technical Implementation and Architecture}
Our Retrieval-Augmented Generation (RAG) system employs a two-stage pipeline designed specifically for financial knowledge integration and domain-specific reasoning enhancement. The implementation addresses the unique challenges of professional financial certification evaluation through carefully engineered prompt templates and vector-based retrieval mechanisms.
\subsubsection{Embedding Infrastructure and Database Organization}
The foundation of our RAG system rests upon OpenAI \texttt{text-embedding-3-small} model, which provides high-dimensional semantic representations optimized for retrieval tasks. This embedding model was selected for its balance between computational efficiency and semantic precision, particularly important given the technical nature of financial terminology and concepts. The embedded representations are stored and managed using Chroma, a specialized vector database that enables efficient similarity-based retrieval operations.
Our knowledge base architecture employs a hierarchical organization strategy that mirrors the CFA curriculum structure. Document collections are partitioned by examination level and further subdivided by topic domains, creating focused retrieval environments that enhance precision and reduce computational overhead. This results in separate vector databases for each level-topic combination. This granular organization ensures that retrieval operations access only relevant curriculum materials, significantly improving both accuracy and efficiency.
\subsubsection{Document Processing and Knowledge Base Construction}
Knowledge base construction employs a sophisticated multi-format document processing approach that accommodates diverse source materials including traditional PDF textbooks and structured Markdown content. The system implements intelligent text segmentation strategies that balance chunk granularity with semantic preservation, utilizing overlapping segments to maintain contextual coherence across document boundaries. Each processed segment receives systematic identification through composite addressing schemes that ensure complete document traceability and enable efficient content management operations.

The architecture incorporates incremental update capabilities that distinguish between existing and novel content, supporting both comprehensive database reconstruction and selective content additions without computational redundancy. This design accommodates the dynamic nature of educational materials while maintaining system efficiency. The hierarchical organization strategy creates domain-specific vector collections that mirror the structured nature of professional certification curricula, resulting in focused search spaces that enhance retrieval precision while optimizing computational resource utilization during similarity matching operations.

\subsubsection{Two-Stage Query Processing}
The first stage transforms complex CFA questions into targeted retrieval queries through structured prompt engineering. Rather than using raw questions directly, the system generates two components: a concise 50-word summary and 5-10 domain-specific keywords. This approach addresses the mismatch between question phrasing and curriculum content organization, enabling more precise retrieval by focusing on fundamental concepts rather than superficial linguistic patterns.
The second stage performs semantic similarity searches across topic-specific databases, retrieving exactly five document segments per query. The similarity search employs cosine similarity metrics between query embeddings and stored curriculum segments, ensuring semantic relevance while maintaining manageable context length for language model processing.
\subsubsection{Context Integration and Response Generation}
Retrieved curriculum segments are integrated into reasoning contexts through structured templates that preserve hierarchical relationships between different curriculum sources. The MCQ reasoning prompt employs a multi-step analytical framework guiding models through systematic problem-solving: understanding question context, identifying applicable CFA concepts, analyzing options against financial principles, and providing structured responses with both choice selection and explanatory reasoning.
\section{Empirical Results}

\subsection{Overall Performance}
\begin{table}[ht]
\centering
\small
\begin{tabular}{llccc}
\hline
\textbf{Level} & \textbf{Model} & \textbf{Zero-shot Accuracy} & \textbf{RAG Accuracy} & \textbf{RAG Improvement (\%)} \\
\hline
\multirow{3}{*}{Level 1} 
 & GPT-4o      & 78.56\% & 79.44\% & +0.89 \\
 & GPT-o1 & \textbf{94.78}\% & \textbf{94.78}\% & +0.00 \\
 & o3-mini     & 87.56\% & 88.33\% & +0.78 \\
\hline
\multirow{3}{*}{Level 2} 
 & GPT-4o      & 59.55\% & 60.45\% & +0.91 \\
 & GPT-o1  & \textbf{89.32}\% & \textbf{91.36}\% & +2.05 \\
 & o3-mini     & 79.77\% & 84.32\% & +4.55 \\
\hline
\multirow{3}{*}{Level 3} 
 & GPT-4o      & 64.09\% & 68.64\% & +4.55 \\
 & GPT-o1  & \textbf{79.09}\% & \textbf{87.73}\% & +8.64 \\
 & o3-mini     & 70.91\% & 76.36\% & +5.45 \\
\hline
\end{tabular}
\caption{Accuracy Comparison and RAG Improvement over Zero-Shot}
\label{tab:overall_accuracy}
\end{table}
Table \ref{tab:overall_accuracy} presents the overall performance of three LLMs across all CFA levels under both zero-shot and RAG-enhanced conditions. \texttt{GPT-o1} consistently achieves the highest accuracy across all levels, demonstrating strong financial reasoning capabilities with zero-shot accuracies of 94.78\%, 89.32\%, and 79.09\% for Levels I, II, and III, respectively. \texttt{o3-mini} shows competitive performance, particularly at Level I (87.56\%), while \texttt{GPT-4o} exhibits more variable results, with notably lower performance on Level II (59.55\%).

Overall, the data indicates that while high-performing models like \texttt{GPT-o1} already excel in zero-shot settings, RAG can further enhance performance, especially for complex financial analysis and for smaller models that benefit more from contextual augmentation.

\textbf{Zero-shot.} In the zero-shot scenario, \texttt{GPT-o1} consistently outperforms the other models across all three levels, showing high potential of passing three levels of MCQs. \texttt{o3-mini} performs reasonably well, while \texttt{GPT-4o} lags behind other models. All models struggle more with level II and III, highlighting the difficulties in handling long context and performing complex reasoning.

\textbf{RAG pipeline.} When supplemented with RAG, all models exhibit performance improvements, although the magnitude of improvement varies across levels. At Level I, where questions are shorter and more straightforward, the improvements of RAG are minimal. This may suggest that the models' built-in knowledge is sufficient for foundational tasks, and that it is trickier for models to extract effective RAG queries with short questions. On the other hand, RAG leads to the most substantial gains at Level 3, where contextual understanding is crucial and more domain-specific knowledge is tested. For instance, the accuracy of \texttt{GPT-o1} is improved by 1.99\% and 8.64\% on Level II and Level III questions, respectively. It indicates that \texttt{GPT-o1} is potentially better at understanding long contexts and utilizing a wide range of sources. \texttt{GPT-4o} also shows notable RAG improvement in Level II and III, while \texttt{o3-mini} demonstrates consistent gains in RAG at 5-6\%.

\subsection{Performance by CFA Level}
The results reveal distinct performance patterns across CFA difficulty levels. Level I represents the most accessible tier, where all models achieve relatively high accuracy, with \texttt{GPT-o1} and \texttt{o3-mini} both exceeding 87\%. The straightforward, foundational nature of Level I questions allows models to leverage their pre-trained financial knowledge effectively.

Level II presents significantly greater challenges, with accuracy dropping substantially across all models. \texttt{GPT-4o} struggles particularly at this level (59.55\%), while \texttt{GPT-o1} maintains strong performance (89.32\%). The increased complexity of case-based scenarios and quantitative analysis demands more sophisticated reasoning capabilities.

Level III shows the most complex pattern, where performance varies significantly by model sophistication. While \texttt{GPT-o1} achieves 79.09\% accuracy, the integration of portfolio management concepts and advanced analytical frameworks proves challenging even for the most capable models. All models demonstrate increased difficulty with the highest level of CFA content, highlighting the progressive complexity of professional financial certification requirements.

\subsection{Performance by Topic}

\begin{table}[ht]
\centering
\scriptsize
\begin{tabular}{lcccccc}
\toprule
\multirow{2}{*}{\textbf{Topic}} & \multicolumn{2}{c}{\textbf{GPT-4o}} & \multicolumn{2}{c}{\textbf{GPT-o1}} & \multicolumn{2}{c}{\textbf{o3-mini}} \\
\cmidrule(lr){2-3} \cmidrule(lr){4-5} \cmidrule(lr){6-7}
 & ZS & RAG & ZS & RAG & ZS & RAG \\
\midrule
Alternative Investments & 81.54\% & 86.15\% & 96.92\% & 98.46\% & 90.77\% & 89.23\% \\
Corporate Issuers       & 84.06\% & 86.96\% & 95.65\% & 91.30\% & 88.41\% & 89.86\% \\
Derivatives             & 81.43\% & 82.86\% & 97.14\% & 94.29\% & 90.00\% & 94.29\% \\
Economics               & 86.36\% & 89.39\% & 95.45\% & 93.94\% & 89.39\% & 87.88\% \\
Equity Investments      & 78.26\% & 79.13\% & 94.78\% & 97.39\% & 93.91\% & 96.52\% \\
Ethics                  & 79.31\% & 79.31\% & 98.28\% & 98.28\% & 93.10\% & 93.10\% \\
Fixed Income            & 82.14\% & 82.14\% & 94.05\% & 94.05\% & 91.67\% & 94.05\% \\
Financial Statement Analysis & 84.62\% & 84.62\% & 96.15\% & 96.15\% & 93.59\% & 93.59\% \\
Portfolio Management    & 68.18\% & 72.73\% & 93.18\% & 93.18\% & 88.64\% & 93.18\% \\
Quantitative Methods    & 86.96\% & 86.96\% & 92.75\% & 92.75\% & 88.41\% & 89.86\% \\
\bottomrule
\end{tabular}
\caption{Accuracy per Topic on Level I}
\label{tab:level1_accuracy}
\end{table}

\begin{table}[ht]
\centering
\scriptsize
\begin{tabular}{lcccccc}
\toprule
\multirow{2}{*}{\textbf{Topic}} & \multicolumn{2}{c}{\textbf{GPT-4o}} & \multicolumn{2}{c}{\textbf{GPT-o1}} & \multicolumn{2}{c}{\textbf{o3-mini}} \\
\cmidrule(lr){2-3} \cmidrule(lr){4-5} \cmidrule(lr){6-7}
 & ZS & RAG & ZS & RAG & ZS & RAG \\
\midrule
Alternative Investments       & 40.63\% & 53.13\% & 87.50\% & 90.63\% & 68.75\% & 84.38\% \\
Corporate Issuers             & 64.29\% & 60.71\% & 96.43\% & 100.00\% & 96.43\% & 100.00\% \\
Derivatives                   & 50.00\% & 50.00\% & 89.29\% & 85.71\% & 78.57\% & 75.00\% \\
Economics                     & 78.57\% & 71.43\% & 92.86\% & 96.43\% & 89.29\% & 89.29\% \\
Equity Investments            & 51.56\% & 50.00\% & 87.50\% & 93.75\% & 81.25\% & 90.63\% \\
Ethics                        & 70.00\% & 70.00\% & 88.33\% & 85.00\% & 58.33\% & 58.33\% \\
Financial Statement Analysis  & 53.33\% & 55.00\% & 88.33\% & 93.33\% & 83.33\% & 96.67\% \\
Fixed Income                  & 60.71\% & 64.29\% & 83.93\% & 85.71\% & 83.93\% & 82.14\% \\
Portfolio Management          & 62.50\% & 66.07\% & 94.64\% & 96.43\% & 83.93\% & 85.71\% \\
Quantitative Methods          & 67.86\% & 64.29\% & 89.29\% & 89.29\% & 85.71\% & 89.29\% \\
\bottomrule
\end{tabular}
\caption{Accuracy per Topic on Level II}
\label{tab:level2_accuracy}
\end{table}

\begin{table}[ht]
\centering
\scriptsize
\begin{tabular}{lcccccc}
\toprule
\multirow{2}{*}{\textbf{Topic}} & \multicolumn{2}{c}{\textbf{GPT-4o}} & \multicolumn{2}{c}{\textbf{GPT-o1}} & \multicolumn{2}{c}{\textbf{o3-mini}} \\
\cmidrule(lr){2-3} \cmidrule(lr){4-5} \cmidrule(lr){6-7}
 & ZS & RAG & ZS & RAG & ZS & RAG \\
\midrule
Alternative Investments & 100.00\% & 100.00\% & 100.00\% & 75.00\% & 100.00\% & 100.00\% \\
Derivatives             & 40.00\% & 45.00\% & 85.00\% & 90.00\% & 75.00\% & 75.00\% \\
Economics               & 72.73\% & 72.73\% & 72.73\% & 90.91\% & 86.36\% & 86.36\% \\
Equity Investments      & 68.75\% & 62.50\% & 75.00\% & 68.75\% & 75.00\% & 62.50\% \\
Ethics                  & 71.43\% & 71.43\% & 76.19\% & 78.57\% & 61.90\% & 69.05\% \\
Fixed Income            & 69.44\% & 86.11\% & 86.11\% & 94.44\% & 77.78\% & 83.33\% \\
Portfolio Management    & 58.75\% & 63.75\% & 77.50\% & 92.50\% & 65.00\% & 76.25\% \\
\bottomrule
\end{tabular}
\caption{Accuracy per Topic on Level III}
\label{tab:level3_accuracy}
\end{table}

The topic-level analysis across Tables \ref{tab:level1_accuracy}–\ref{tab:level3_accuracy} reveals distinct strengths and weaknesses for each model. 

\textbf{Model-specific topic performance:} \texttt{GPT-4o} shows particular weakness in quantitative and analytical topics, especially Alternative Investments, Derivatives, and Equity Investments at Level II, where performance drops significantly (40.63\%, 50.00\%, and 51.56\% respectively). These areas demand quantitative modeling and numerical accuracy, highlighting \texttt{GPT-4o}'s limitations in complex mathematical reasoning. \texttt{GPT-o1}, by contrast, delivers consistently high accuracy on core Level I subjects and proves versatile across higher-level material, excelling in Level II Corporate Issuers and Portfolio Management, as well as Level III Derivatives and Fixed Income. However, even this most capable model shows one notable shortcoming in Level III Equity Investments, where the RAG pipeline offers only modest help. \texttt{o3-mini} demonstrates balanced performance across most topics, matching \texttt{GPT-o1}'s solid performance on Level I topics and showing strong reasoning and computational ability in Level II Corporate Issuers and Economics, and Level III Economics and Fixed Income. However, it struggles markedly with Ethics at Levels II and III (58.33\% at both levels), suggesting limitations either in its retrieval of relevant case precedents or in the depth of its ethical reasoning heuristics.

\textbf{Topic complexity patterns:} Certain topics prove consistently challenging across models, revealing inherent complexities in specific CFA domains. Portfolio Management shows relatively lower performance for \texttt{GPT-4o} across all levels, while Alternative Investments at Level II presents difficulties for both \texttt{GPT-4o} and \texttt{o3-mini}. These patterns suggest that topics requiring integration of multiple concepts and real-world application present greater challenges than foundational knowledge areas. The complexity varies not only by mathematical requirements but also by the need for contextual understanding and practical application. Overall, these results demonstrate that \texttt{GPT-o1} emerges as the most adaptable across domains, \texttt{GPT-4o} remains constrained by math-heavy topics, and \texttt{o3-mini} demonstrates broad competence but notable gaps in specialized areas like ethical reasoning.

\textbf{Zero-shot.} Zero-shot accuracy confirms that \texttt{GPT-o1} consistently outperforms \texttt{GPT-4o} and \texttt{o3-mini} across all levels and nearly all topics. In contrast, \texttt{GPT-4o} shows more fluctuation, with notably lower Zero-Shot performance in Level II topics like Equity Investments (52.08\%) and Derivatives (54.17\%).  The smaller \texttt{o3-mini} model maintains competitive accuracy in Level I, particularly in topics like Quantitative Methods and Ethics, but tends to trail \texttt{GPT-o1} in more complex levels. 

\textbf{RAG pipeline.} While RAG improves the overall accuracy for all models and all levels, there is strong variation among topics. For example, RAG helps to improve level II and III topics Fixed Income, and Portfolio Management substantially (2.27\%-16.67\%). However, it seems to lead to confusion in LLMs performance in Level III Equity Investments. 

\subsection{RAG Impact Analysis}

The RAG pipeline demonstrates variable effectiveness across different contexts, with several key patterns emerging from the data. Level-dependent improvements show that RAG provides minimal benefits at Level I (improvements of 0.89\%, 0.00\%, and 0.78\% for \texttt{GPT-4o}, \texttt{GPT-o1}, and \texttt{o3-mini} respectively), where models' built-in knowledge appears sufficient for foundational questions. However, RAG impact increases substantially at higher levels, with Level III showing the most significant gains: 4.55\%, 8.64\%, and 5.45\% improvements respectively.

Topic-specific RAG effectiveness varies considerably. RAG provides substantial benefits for knowledge-intensive topics like Alternative Investments at Level II, where \texttt{GPT-4o} improves from 40.63\% to 53.13\%, and Fixed Income at Level III, where improvements range from 2.27\% to 16.67\%. However, RAG occasionally reduces performance, as seen in Level III Equity Investments, where retrieval appears to introduce confusing or irrelevant information.

Model-dependent RAG utilization reveals that \texttt{GPT-o1} benefits most from RAG at higher complexity levels, with its Level III improvement of 8.64\% suggesting superior ability to integrate retrieved information with reasoning processes. \texttt{o3-mini} shows consistent moderate gains (4.55\%-5.45\%), while \texttt{GPT-4o} demonstrates variable RAG effectiveness, performing better when retrieval addresses specific knowledge gaps but struggling when additional context increases complexity.

Based on the estimated passing criteria, \texttt{GPT-o1} and \texttt{o3-mini} can pass CFA Level I, Level II, and the MCQ part of Level III, with or without the help of RAG pipeline. The \texttt{GPT-4o} model alone can pass Level II, but struggles with Level II and III, especially in certain topics like Alternative Investments in Level II and Derivatives in Level III. With RAG, \texttt{GPT-4o} can reach the passing standard of Level II exam, as the proposed RAG pipeline significantly improves its accuracy on weak topics, such as Alternative Investments in Level II (from 40.63\% to 53.13\%).

\section{Analysis and Discussion}

This section offers a comprehensive assessment of each model’s capacity for financial analysis. It identifies the primary sources of error, evaluates the influence of text readability, and quantifies the share of mistakes arising from conceptual versus calculation-based questions.

\subsection{Error Type}
We obtain the LLMs' responses and analyze the explanations given by these LLMs. The errors are categorized into four types, following the definitions in \cite{callanan2023gptmodelsfinancialanalysts}:
\begin{enumerate}
    \item \textbf{Knowledge errors:} Where the model lacks or holds wrong knowledge. This includes an incorrect understanding of some concept, not knowing the relationship between concepts, or using an incorrect formula.
    \item \textbf{Reasoning errors:} When the model had all the correct knowledge, but misinterpreted the question, made incorrect intermediate deductions, or hallucinated information that was not actually present.
    \item \textbf{Calculation errors:} Incorrect calculation (using a correct formula), or failing to accurately compare or convert results.
    \item \textbf{Inconsistency errors:} When the model’s thinking is entirely correct, yet it chooses the wrong answer.
\end{enumerate}

We focus on the RAG pipeline to probe the intrinsic capabilities of GPT models without mixing pure reasoning with lack of context.As shown in Table \ref{tab:error_distribution}, knowledge errors dominate across all models and levels, reflecting the specialized nature of CFA content that draws on specific accounting rules, valuation conventions, and proprietary ethical codes. The prevalence of knowledge-based failures underscores the critical importance of comprehensive domain knowledge in financial reasoning tasks.

Model-specific error patterns reveal distinct behavioral characteristics. \texttt{GPT-4o} exhibits the most diverse error profile, being the only model to produce significant inconsistency errors across all levels (29.73\% at Level I, 26.44\% at Level II, and 14.49\% at Level III). It also demonstrates higher susceptibility to calculation errors, particularly at Level I (18.92\%), indicating challenges with both numerical accuracy and answer selection consistency.

\texttt{GPT-o1} shows a dramatically different pattern, with knowledge errors comprising the majority of its mistakes (40.43\% at Level I, 78.95\% at Level II, and 70.37\% at Level III). Notably, it produces virtually no inconsistency errors at Levels II and III, and minimal calculation errors across all levels, suggesting strong logical coherence once relevant knowledge is available. The increasing proportion of knowledge errors at higher levels reflects the greater domain complexity of advanced CFA topics.

\texttt{o3-mini} demonstrates a unique profile with reasoning errors being particularly prominent at Level I (57.14\%), though knowledge errors dominate at higher levels (82.35\% at Level II, 57.69\% at Level III). Like \texttt{GPT-o1}, it produces no inconsistency errors at Levels II and III, indicating reliable answer selection mechanisms despite its lighter architecture.

Level-dependent patterns show increasing knowledge error prevalence as exam complexity rises. For \texttt{GPT-o1} and \texttt{o3-mini}, knowledge errors constitute over 70\% of mistakes at Levels II and III, compared to approximately 40\% at Level I. This pattern reflects the progressive specialization of CFA curriculum, where advanced topics require deep domain expertise less likely to be captured in general training data.

The near-elimination of inconsistency errors in newer models (\texttt{GPT-o1} and \texttt{o3-mini}) compared to \texttt{GPT-4o} suggests architectural improvements in answer selection mechanisms. Similarly, the reduced calculation error rates in advanced models indicate enhanced numerical processing capabilities, though complex mathematical operations remain challenging across all architectures.

These findings suggest that targeted knowledge augmentation through improved RAG systems may yield greater performance gains than simply upgrading to larger models, particularly given the dominance of knowledge-based errors across all evaluated architectures.

\begin{table}[ht]
\centering
\scriptsize
\begin{tabular}{llrrrrr}
\toprule
\textbf{Level} & \textbf{Model}    & \textbf{Knowledge}        & \textbf{Reasoning}        & \textbf{Calculation}       & \textbf{Inconsistency}      & \textbf{Total}            \\
\midrule
\multirow{3}{*}{1} & GPT-4o   & 36 (19.46\%)      & 59 (31.89\%)      & 35 (18.92\%)       & 55 (29.73\%)        & 185 (100.00\%)   \\
                   & GPT-o1   & 19 (40.43\%)      & 23 (48.94\%)      & 2  (4.26\%)        & 3  (6.38\%)         & 47  (100.00\%)   \\
                   & o3-mini  & 38 (36.19\%)      & 60 (57.14\%)      & 4  (3.81\%)        & 3  (2.86\%)         & 105 (100.00\%)   \\
\midrule
\multirow{3}{*}{2} & GPT-4o   & 78 (44.83\%)      & 37 (21.26\%)      & 13 (7.47\%)        & 46 (26.44\%)        & 174 (100.00\%)   \\
                   & GPT-o1   & 30 (78.95\%)      & 5  (13.16\%)      & 3  (7.89\%)        & 0  (0.00\%)         & 38  (100.00\%)   \\
                   & o3-mini  & 56 (82.35\%)      & 12 (17.65\%)      & 0  (0.00\%)        & 0  (0.00\%)         & 68  (100.00\%)   \\
\midrule
\multirow{3}{*}{3} & GPT-4o   & 51 (73.91\%)      & 5  (7.25\%)       & 3  (4.35\%)        & 10 (14.49\%)        & 69  (100.00\%)   \\
                   & GPT-o1   & 19 (70.37\%)      & 8  (29.63\%)      & 0  (0.00\%)        & 0  (0.00\%)         & 27  (100.00\%)   \\
                   & o3-mini  & 30 (57.69\%)      & 19 (36.54\%)      & 3  (5.77\%)        & 0  (0.00\%)         & 52  (100.00\%)   \\
\bottomrule
\end{tabular}
\caption{Error type distribution by level and model (RAG pipeline)}
\label{tab:error_distribution}
\end{table}

\subsection{Readability}

\begin{figure}[ht]
    \centering
    \includegraphics[width=\textwidth]{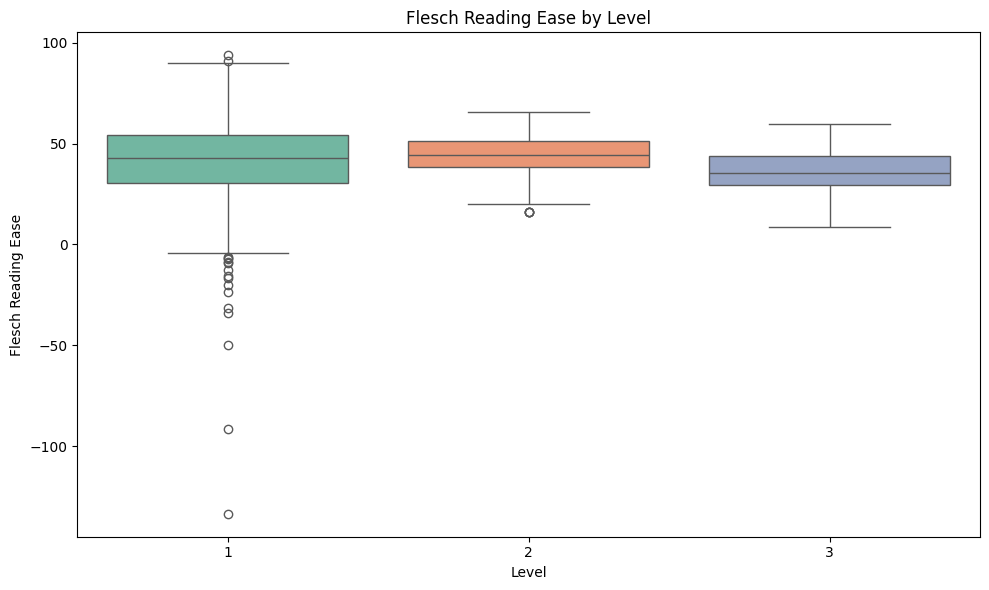}
    \caption{Flesch Reading Ease of CFA questions}
    \label{fig:readability}
\end{figure}

\begin{figure}[ht]
    \centering
    \includegraphics[width=\textwidth]{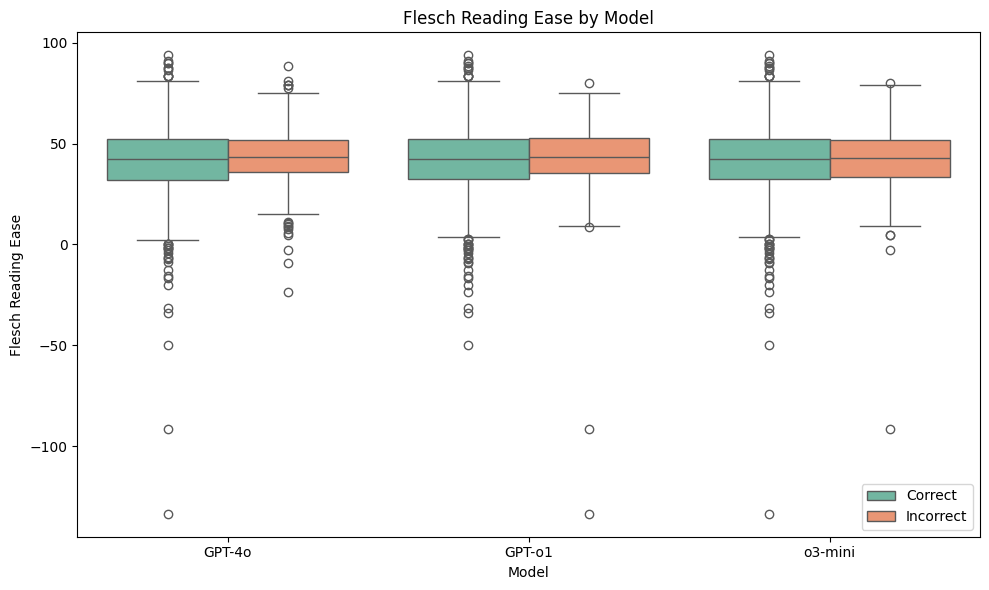}
    \caption{Flesch Reading Ease by model}
    \label{fig:readability_by_model}
\end{figure}

Since the mock questions contain long case description, multiple tables, and numbers, it is helpful to assess how difficult the questions are to understand, and how LLMs are affected by the readability. To quantify textual difficulty, we compute Flesch Reading Ease (FRE) for every question (\cite{reading_ease}). FRE converts sentence length and syllable density into a score that ranges from $<$ 0 (“very confusing”) to 100 (“very easy”). 

Figure \ref{fig:readability} plots FRE by exam level. CFA Institute materials are written for university students and professions from all backgrounds, the scores in the 30–60 band are within expectation, roughly equivalent to undergraduate-level prose. Level 1 items show the widest spread and include a long left tail of technical disclosures (FRE $<$ 0). Levels 2 and 3 cluster more tightly around the mid-40s, indicating moderately difficult but uniformly edited narratives. In other words, while Level I occasionally embed dense words and complex sentences, Level II and III achieve a consistent editorial tonality that neither simplifies nor substantially complicates the reading task.

Figure \ref{fig:readability_by_model} shows overlaying FRE distributions for correct and incorrect responses across the three GPT models. The medians for correct and incorrect subsets differ minimally, which confirms that the gap of readability is not significant. We can thus conclude that passages written in plainer language are no easier for the models, nor are densely worded questions systematically harder.

The absence of a readability effect carries two implications. Firstly, modern GPT models already exceed the basic reading-comprehension threshold required for financial exam questions. Accuracy bottlenecks should arise elsewhere, primarily in domain knowledge retrieval and quantitative reasoning. Secondly, curating simpler question wording is unlikely to boost performance. Instead, we should invest in retrieval pipelines that surface relevant concepts such as authoritative accounting rules, valuation formulas, and ethical guidelines, and in numeric-verification techniques to catch arithmetic errors.

\subsection{Question Type}

\begin{table}[ht]
\centering
\scriptsize
\begin{tabular}{lcccccc}
\toprule
\multirow{2}{*}{\textbf{Topic}} 
  & \multicolumn{2}{c}{\textbf{Level 1}} 
  & \multicolumn{2}{c}{\textbf{Level 2}} 
  & \multicolumn{2}{c}{\textbf{Level 3}} \\
\cmidrule(lr){2-3} \cmidrule(lr){4-5} \cmidrule(lr){6-7}
  & Calculation & Conceptual 
  & Calculation & Conceptual 
  & Calculation & Conceptual \\
\midrule
Alternative \\Investments      &  7 &  58 & 15 & 17 &  0 &  4 \\
Corporate Issuers             &  6 &  63 & 16 & 12 &  0 &  0 \\
Derivatives                   & 11 &  59 & 15 & 13 & 14 &  6 \\
Economics                     &  6 &  60 & 11 & 17 &  3 & 19 \\
Equity Investments            & 30 &  85 & 40 & 24 &  0 & 16 \\
Ethics                        &  0 & 135 &  1 & 59 &  0 & 42 \\
Financial Statement \\Analysis & 27 &  83 & 17 & 43 &  0 &  0 \\
Fixed Income                  & 35 &  75 & 29 & 27 &  6 & 30 \\
Portfolio Management          & 12 &  78 & 20 & 36 & 17 & 63 \\
Quantitative Methods          & 28 &  42 &  8 & 20 &  0 &  0 \\
\midrule
\textbf{Total}                &162 & 738 &172 &268 & 40 &180 \\
\bottomrule
\end{tabular}
\caption{Overview of question type}
\label{tab:question_type_overview}
\end{table}

\begin{table}[ht]
\centering
\scriptsize
\begin{tabular}{lcccccc}
\toprule
\multirow{2}{*}{\textbf{Model}} 
  & \multicolumn{2}{c}{\textbf{Correct Answers}} 
  & \multicolumn{2}{c}{\textbf{Incorrect Answers}} 
  & \multirow{2}{*}{\textbf{Total}} \\
\cmidrule(lr){2-3} \cmidrule(lr){4-5}
  & Calculation  & Conceptual 
  & Calculation  & Conceptual & \\
\midrule
GPT-4o    & 167 (10.71\%) & 943 (60.45\%) & 207 (13.27\%) & 243 (15.58\%) & 1560 \\
GPT-o1    & 346 (22.18\%) & 1074 (68.85\%) &  28 (1.79\%)  & 112 (7.18\%)  & 1560 \\
o3-mini   & 331 (21.22\%) & 964 (61.79\%)  &  43 (2.76\%)  & 222 (14.23\%) & 1560 \\
\bottomrule
\end{tabular}
\caption{Accuracy and error rate by question type (zero-shot)}
\label{tab:question_type_ZS}
\end{table}

\begin{table}[H]
\centering
\scriptsize
\begin{tabular}{lcccccc}
\toprule
\multirow{2}{*}{\textbf{Model}} 
  & \multicolumn{2}{c}{\textbf{Correct Answers}} 
  & \multicolumn{2}{c}{\textbf{Incorrect Answers}} 
  & \multirow{2}{*}{\textbf{Total}} \\
\cmidrule(lr){2-3} \cmidrule(lr){4-5}
  & Calculation & Conceptual
  & Calculation & Conceptual & \\
\midrule
GPT-4o   & 156 (10.00\%) & 976 (62.56\%) & 218 (13.97\%) & 210 (13.46\%) & 1560 \\
GPT-o1   & 349 (22.37\%) & 1099 (70.45\%) &  25 (1.60\%)  &  87 (5.58\%)  & 1560 \\
o3-mini  & 342 (21.92\%) &  992 (63.59\%) &  32 (2.05\%)  & 194 (12.44\%) & 1560 \\
\bottomrule
\end{tabular}
\caption{Accuracy and error rate by question type (RAG)}
\label{tab:question_type_rag}
\end{table}

The distribution of types of errors suggest that LLMs' performance varies between reasoning and calculation tasks. Therefore, each MCQ was classified as \textbf{conceptual} or \textbf{calculation} to disentangle numerical ability from pure financial understanding. This split is important: an LLM could in principle excel at narrative recall yet fail disastrously at arithmetic, or vice-versa. We prompted GPT model with a short rubric and asked for a binary tag. The overview of the dataset is given in Table \ref{tab:question_type_overview}. Consistent with the CFA curriculum’s emphasis on integrated knowledge and scenario analysis, roughly four-fifths of the questions are conceptual. The remaining fifth required explicit computations, either requiring numerical as final results or in intermediate steps. The count and percentage of correct and incorrect answers are provided in Table \ref{tab:question_type_ZS} for zero-shot and Table \ref{tab:question_type_rag} for RAG pipeline.

\textbf{Zero-shot.} Table \ref{tab:question_type_ZS} shows that GPT-o1 achieves the highest overall accuracy , followed by o3-mini and the older GPT-4o, in all question types. Across the board, models answer conceptual questions more reliably than calculation ones, but the magnitude of the gap varies sharply. For GPT-4o, calculation accuracy is only 44.7 \% (167/374), whereas conceptual items are answered correctly 79.5 \% of the time. Nearly half of its wrong answers stem from mis-computations, although calculation questions contribute to only about one-fifth of total questions. By contrast, GPT-o1 and o3-mini answer more than 88 \% of calculation items correctly, so only about one-fifth (GPT-o1) or one-sixth (o3-mini) of their errors occur in calcualtion questions. In other words, newer models have largely closed the numeric gap that plagued earlier releases.

\textbf{RAG pipeline.} Retrieval-augmented generation (Table \ref{tab:question_type_rag}) lifts conceptual accuracy for all three models, adding 2–3 percentage points across the board, but has almost no effect on calculation accuracy. Consequently, each model’s error rate shifts accordingly. GPT-4o’s mistakes are still evenly split between calculation and conceptual questions, whereas GPT-o1 and o3-mini still see fewer than a quarter of their residual errors coming from arithmetic. This suggests that the RAG pipeline supplies authoritative financial concepts, yet the models still have to execute the math correctly for calculation questions.

Taken together, the split‐out results reveal that model capacity and retrieval serve complementary roles: GPT-o1 and o3-mini already handle arithmetic remarkably well in both settings (about 90 \% accuracy on calculation items, with only 2 \%–3 \% of their total answers wrong for numeric reasons), whereas GPT-4o’s older architecture still falters on math (about 45 \% accuracy) so nearly half of its errors remain calculation-driven even after retrieval. The RAG pipeline boosts conceptual accuracy for every model, because missing IFRS nuances, option‐payoff definitions, and ethical clauses are now surfaced, but leaves calculation accuracy virtually unchanged, underscoring that context alone cannot force correct numeric execution. In short, retrieval augmentation is indispensable for fact-heavy conceptual judgment, while deterministic verification layers or newer instruction-tuned models are required to close the remaining gap in quantitative reasoning.

\section{Practical Implications}
\subsection{Model Selection and Cost-Performance Trade-offs}
Although larger models provide stronger reasoning abilities, they are also priced higher. As of March 2025, the price per 1 million tokens for \texttt{GPT-4o}, \texttt{GPT-o1}, and \texttt{o3-mini} is \$2.50, \$15.00, and \$1.10, respectively. Practitioners need to consider the trade-off between performance and costs when deploying LLMs in financial contexts.

Given the substantial performance differences observed, \texttt{GPT-o1} should be reserved for complex and high-stakes financial analysis where accuracy is paramount, such as regulatory compliance assessments, advanced portfolio management, and client-facing investment recommendations. Despite its premium pricing, the model's consistent 85\%+ accuracy across all CFA levels and effective RAG utilization justify the cost for scenarios where analytical errors could have significant financial or regulatory consequences.

For high-volume, routine financial tasks, \texttt{o3-mini} provides reliable performance at substantially lower costs. Its 87.56\% accuracy on foundational topics and consistent RAG improvements make it suitable for preliminary document analysis, basic financial calculations, and educational applications where moderate accuracy levels are acceptable.
\texttt{GPT-4o} presents challenges for professional financial deployment due to inconsistent performance, particularly its 59.55\% accuracy on Level II questions and notable calculation errors (44.7\% accuracy on quantitative tasks). While RAG can partially address these limitations, practitioners should carefully evaluate whether the intermediate pricing justifies its variable performance.
\subsection{RAG Implementation Strategy}
Regarding retrieval augmentation, RAG provides substantial benefits for knowledge-intensive scenarios and complex case-based analysis, with improvements of up to 8.64\% observed at higher complexity levels. However, RAG offers minimal benefits for foundational questions and may occasionally introduce confusion. Practitioners should implement RAG selectively for topics requiring integration of multiple concepts, frequently updated regulations, or institution-specific policies.
\subsection{LLM Financial Applications Based on CFA Performance}
The CFA evaluation results reveal specific strengths that translate to practical financial applications. Models' strong performance on Ethics topics (93\%+ accuracy for advanced models) suggests reliable deployment for compliance monitoring and regulatory interpretation, where consistent application of professional standards is critical. The superior accuracy on foundational Level I topics (87\%+ for \texttt{o3-mini}) supports educational and training applications, including CFA exam preparation, continuing education programs, and junior analyst development.

For document analysis and research support, the models demonstrate capability in processing financial statements, earnings reports, and regulatory filings, particularly when augmented with RAG systems. However, the significant performance drop at Level II (GPT-4o: 59.55\%) indicates limitations for complex case-based analysis requiring integration of multiple financial concepts.

Portfolio management and investment analysis applications should be approached cautiously given the calculation error rates observed. While \texttt{GPT-o1} shows promise for qualitative analysis and conceptual framework application, the quantitative limitations necessitate independent verification systems for numerical outputs. The variable performance across CFA topics suggests matching model deployment to task complexity: routine financial calculations and client communication for cost-effective models, while reserving advanced reasoning capabilities for sophisticated analytical tasks requiring deep domain knowledge integration.

\subsection{Quality Assurance and Deployment Recommendations}
The dominance of knowledge errors (61.68\% of mistakes) across all models highlights the critical importance of maintaining high-quality, up-to-date knowledge bases for RAG systems. Additionally, the significant calculation error rates, particularly for \texttt{GPT-4o}, suggest implementing deterministic verification mechanisms for quantitative outputs in production environments.

Organizations should adopt a tiered deployment strategy: \texttt{GPT-o1} with comprehensive RAG for high-stakes analysis, \texttt{o3-mini} with selective RAG for routine tasks, and robust human oversight for all consequential financial decisions regardless of model choice.

\section{Contributions}
This study makes several significant contributions to the intersection of artificial intelligence and financial technology, providing both methodological innovations and practical insights for the deployment of LLMs in professional financial contexts.

\subsection{Comprehensive Benchmarking of Financial Reasoning in LLMs}
We conduct the first large-scale, systematic evaluation of financial reasoning capabilities in state-of-the-art LLMs (GPT-4o, GPT-o1, and o3-mini) using 1,560 official CFA mock questions spanning all three exam levels. This granular benchmarking reveals model strengths and weaknesses across a diverse range of cognitive tasks, from quantitative calculations to ethical reasoning and portfolio theory.
\subsection{Novel Domain Reasoning RAG Pipeline for Financial Knowledge Integration}
We develop a novel domain reasoning Retrieval-Augmented Generation (RAG) pipeline that integrates raw CFA curriculum documents into the LLM workflow. Our pipeline dynamically retrieves and injects relevant contextual knowledge to enhance model comprehension—without requiring fine-tuning or curated knowledge graphs—demonstrating a practical and scalable strategy for domain adaptation.
\subsection{In-Depth Error Taxonomy and Diagnostic Performance Analysis}
We conduct systematic error categorization that classifies model failures across four distinct dimensions—knowledge gaps, reasoning errors, calculation inaccuracies, and response inconsistencies—providing empirical foundations for model improvement, training optimization, and reliable deployment in professional financial contexts.
\subsection{Practical Guidance for LLM Adoption in Financial Services}
We establish evidence-based deployment frameworks that enable financial institutions to strategically integrate LLMs while managing operational risk and regulatory compliance. Our analysis provides actionable guidelines for model selection, cost-performance optimization, and risk management protocols, distinguishing appropriate use cases across different accuracy requirements and computational budgets. We demonstrate that targeted knowledge augmentation delivers superior performance gains compared to uniform model scaling, while our systematic error analysis informs verification protocols essential for high-stakes financial applications. By quantifying LLM capabilities across granular topics and difficulty levels, and by illustrating a lightweight augmentation pipeline, we provide a practical foundation for financial professionals, educators, and AI engineers seeking to integrate LLMs into decision-support, advisory, or educational systems.
\section{Conclusion}

Benchmarking the performance of LLMs on financial tasks and identifying their strengths and weaknesses is essential for informed and effective deployment in practical applications. In this study, we systematically evaluate the financial capabilities of recent LLMs, including \texttt{GPT-4o}, \texttt{GPT-o1}, and \texttt{o3-mini}, using multiple-choice questions from CFA mock exams. Our evaluation includes both a zero-shot prompting setting and a domain reasoning RAG pipeline, which supplements model responses with relevant content from the official CFA curriculum. The RAG system guides models to generate retrieval queries based on a brief summary and keyword list derived from the question, allowing them to answer the questions while referring authoritative, context-specific materials. Additionally, we conduct a detailed error analysis, categorizing model failures into four types: Knowledge, Reasoning, Calculation, and Inconsistency.

The results indicate that all models exhibit reduced performance on higher-level CFA questions, which tend to involve longer case contexts, multiple tables, and deeper domain knowledge. The RAG pipeline consistently improves accuracy, especially in more complex scenarios. However, the magnitude of improvement varies by topic. In some cases, overly specific or biased retrievals can hinder performance, highlighting the need for more refined retrieval strategies. Among the models evaluated, \texttt{GPT-o1} consistently outperforms others across all levels and topics, demonstrating both strong zero-shot capabilities and effective use of retrieved information. \texttt{o3-mini}, while more lightweight, delivers stable and cost-efficient performance and shows consistent gains when supplemented with RAG. 

Error forensics reveal that nearly two-thirds of residual mistakes stem from missing or misremembered curriculum facts, whereas arithmetic slips and option mismatches account for less than 15 \% combined. Because passage readability exerts negligible influence on accuracy, future gains will come from (1) curating concise, high-coverage corpora for targeted retrieval, and (2) adding deterministic numerical checks rather than rewriting question prose. Meanwhile, cost analysis suggests a tiered deployment strategy: use GPT-o1 for high-stakes or highly contextual tasks, and delegate routine, high-volume queries to o3-mini .

By showing that smart retrieval and modest verification layers can yield larger practical dividends than upgrading to ever-larger models, our findings re-orient future research toward corpus design, retrieval fidelity, and mathematical ability enhancement that will be essential to shape LLMs as trusted agents in consequential financial decision-making.

\newpage

\bibliography{reference}

\newpage

\appendix

\section{RAG Improvements by Topic}

\begin{figure}[H]
    \centering
    \includegraphics[width=\textwidth]{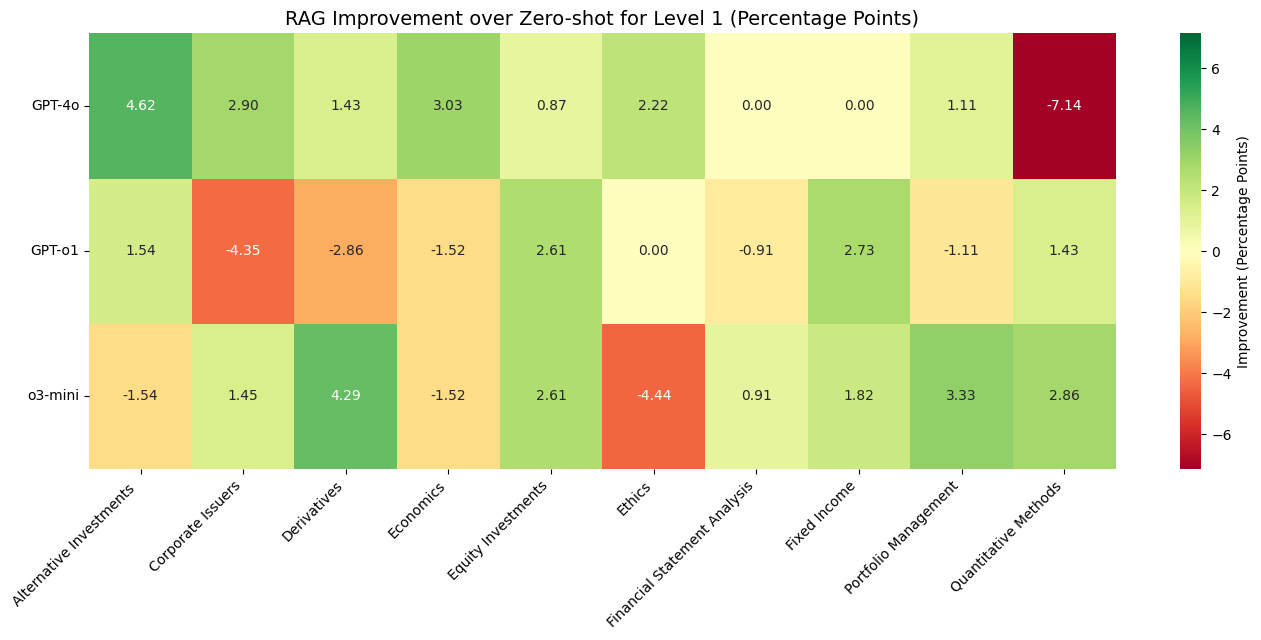}
    \caption{RAG Improvements by Topic on Level I}
    \label{fig:rag_improvement_level1}
\end{figure}

\begin{figure}[H]
    \centering
    \includegraphics[width=\textwidth]{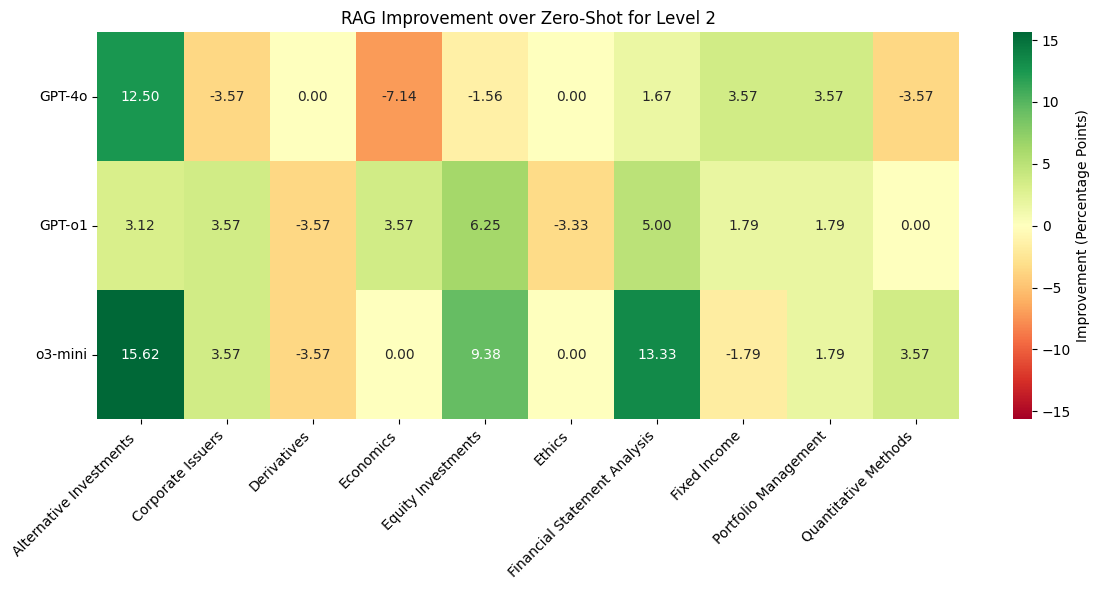}
    \caption{RAG Improvements by Topic on Level I}
    \label{fig:rag_improvement_level2}
\end{figure}

\begin{figure}[H]
    \centering
    \includegraphics[width=\textwidth]{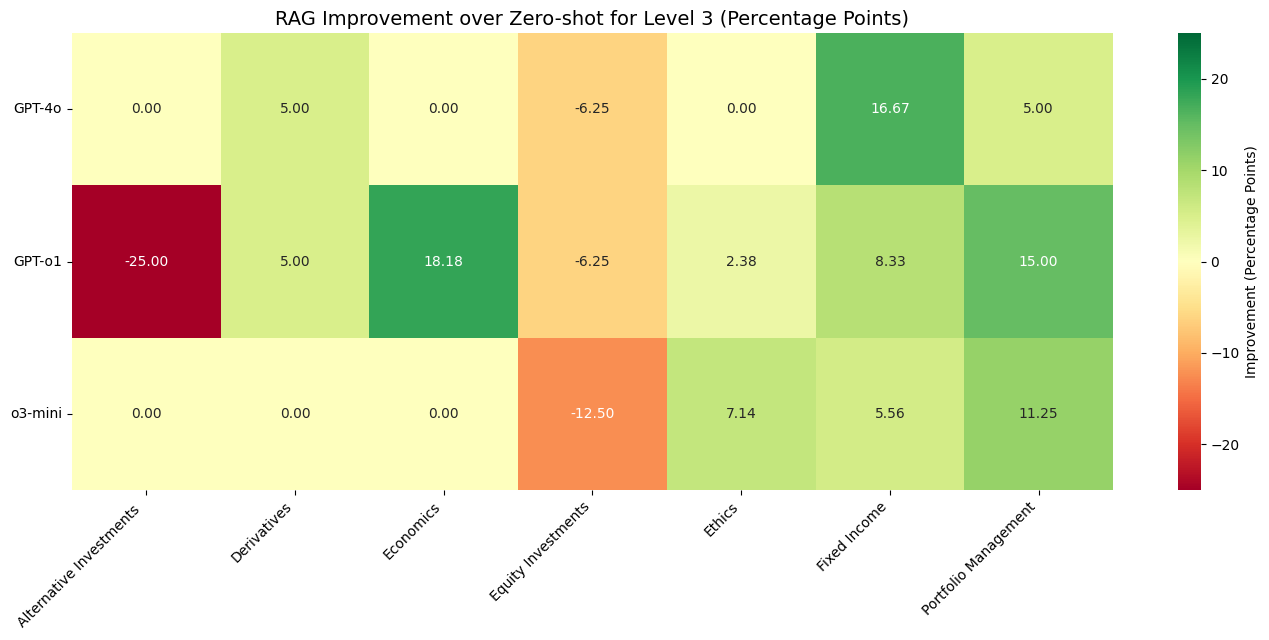}
    \caption{RAG Improvements by Topic on Level I}
    \label{fig:rag_improvement_level3}
\end{figure}

\end{document}